\documentclass[lettersize,journal]{IEEEtran}
\usepackage{amsmath,amsfonts}
\usepackage{algorithmic}
\usepackage{algorithm}
\usepackage{array}
\usepackage[caption=false,font=normalsize,labelfont=sf,textfont=sf]{subfig}
\usepackage{textcomp}
\usepackage{stfloats}
\usepackage{url}
\usepackage{verbatim}
\usepackage{graphicx}
\usepackage{cite}
\usepackage{color}
\usepackage{times}
\usepackage{amsmath}
\usepackage{amsthm}
\usepackage{booktabs}
\usepackage{setspace}
\usepackage{amssymb}
\usepackage{epstopdf}
\usepackage{booktabs}
\usepackage{float}

\usepackage{multirow}
\usepackage{array}
\usepackage[table]{xcolor}
\usepackage{amsmath}

\begin{document}

\title{
Reliable and Compact Graph Fine-tuning via Graph Sparse Prompting
\author{Bo Jiang, Hao Wu, Beibei Wang, Jin Tang, and Bin Luo}

 \thanks{Bo Jiang, Hao Wu, Beibei Wang, Jin Tang, and Bin Luo are with the School of Computer Science and Technology, Anhui University, Hefei 230601, China (e-mail: jiangbo@ahu.edu.cn)}
}

\markboth{Journal of \LaTeX\ Class Files,~Vol.~14, No.~8, August~2021}%
{Shell \MakeLowercase{\textit{et al.}}: A Sample Article Using IEEEtran.cls for IEEE Journals}

\IEEEpubid{}

\maketitle

\begin{abstract}
Recently, graph prompt learning has garnered
increasing attention in adapting pre-trained GNN models for downstream graph learning tasks. 
However, existing works generally conduct prompting over all
graph elements (e.g., nodes, edges, node attributes, etc.), which is  suboptimal and  obviously redundant. 
To address this issue,
we propose exploiting sparse representation theory for graph prompting and
present Graph Sparse Prompting (GSP). GSP aims to 
adaptively and sparsely select the optimal elements (e.g., certain node attributes)
to achieve compact prompting for downstream tasks. 
Specifically, we propose two kinds of GSP models, termed Graph Sparse Feature Prompting
(GSFP) and Graph Sparse multi-Feature Prompting (GSmFP).
Both GSFP and GSmFP provide a general scheme for tuning any specific pre-trained GNNs that can achieve attribute
selection and compact prompt learning simultaneously. A simple yet
effective algorithm has been designed for solving GSFP and GSmFP models. 
Experiments on 16 widely-used benchmark datasets validate  the effectiveness and advantages of the proposed GSFPs.

\end{abstract}


\section{Introduction}
Graph Neural Networks (GNNs) have become a powerful tool for analyzing graph-structured data, with practical applications in many fields such as social network analysis~\cite{social_networks_gnn}, drug discovery~\cite{drug_discovery_gnn}, fraud detection~\cite{fraud_detection_gnn} and traffic prediction~\cite{traffic_predictio_gnn}.
However, GNNs often rely on supervised learning, which demands large amounts of task-specific labeled data. Acquiring such labeled data is time-consuming and expensive, limiting the practical application of GNNs in many real-world scenarios. 
To reduce dependence on labeled data, researchers have explored `pre-training + fine-tuning' strategy~\cite{pretrain_survey}. 
This strategy first learns useful representations from large-scale unlabeled data to enhance the generalization ability of GNNs and then fine-tunes the pre-trained model on the downstream tasks with a small amount of labeled data.
However, due to the discrepancy  between the pre-training and downstream stages in different graph domains, direct fine-tuning may result in suboptimal performance or even lead to negative transfer~\cite{negative_transfer}.
Also, full fine-tuning the pre-trained model is usually time-consuming and computationally expensive. 

To overcome these issues, graph prompt learning techniques have been developed for fine-tuning pre-trained GNN models. By reformulating the downstream data/task to better align with the pre-training phase, these techniques can reduce the gap between the pre-training and downstream stages~\cite{graph_prompt_survey}. 
Overall, existing graph prompt learning methods can  be mainly categorized into two types, i.e., task aligning and graph data prompting.
For task aligning prompting, some works focus on constructing prompt templates to unify the pre-training objective and downstream task. For example, GPPT~\cite{GPPT} converts the downstream node classification task into the pre-trained link prediction task. GraphPrompt~\cite{GraphPrompt} transforms both link prediction in pre-training and the downstream node/graph classification into general subgraph similarity computation. In work~\cite{All-in-One}, it translates both node-level and edge-level downstream tasks into unified graph-level tasks. 
For graph data prompting, some works aim to conduct prompting directly on the input graph \emph{nodes} or \emph{edges}. 
For graph edge prompting, 
AAGOD\cite{AAGOD} proposes edge-level prompt by adjusting edge weights in the adjacency matrix with a learnable amplifier.
PSP\cite{PSP} adds weighted connection edges between class prototype nodes and original nodes in the input graph as structure prompt.
For graph node prompting,
GraphPrompt \cite{GraphPrompt} multiplies the hidden node features with a learnable prompt vector to assist the readout operation. 
All-in-One~\cite{All-in-One} designs the prompt as a graph composed of a set of prompt nodes and inserts it into the input graph.
GPF~\cite{GPF} and GPF-plus~\cite{GPF} directly add prompt vectors to the input node features.
Other graph prompt designs~\cite{SUPT,GSPF,VNT} can generally be seen as special cases or natural extensions of GPF and All-in-One. 
Some works~\cite{GPF,theoretical_support} have theoretically proven the effectiveness of these prompt methods. 
In this paper, we mainly focus on graph node prompt learning. 

After reviewing the above existing graph prompting methods, we observe that they generally conduct  prompting over all graph elements, such as nodes, edges or node attributes. 
For example, 
VNT~\cite{VNT} sends the prompt nodes and the original node set into the Graph Transformer together to achieve prompting on all nodes.
GPF~\cite{GPF} conducts prompting on all node attributes. 
However,
in many downstream tasks, graphs are usually complex, such as large scale, dense edges or high attribute dimension.  
Therefore, two natural questions emerge: 
\emph{Is it necessary to conduct prompting on all graph elements? What are the
most desirable elements needing to be prompted?}

To address these questions, 
for the first time, we propose a new prompt learning problem for graph fine-tuning, termed Graph Sparse Prompting (GSP). 
Generally speaking, our GSP problem is defined as adaptively and sparsely selecting the
optimal elements to achieve compact  prompting for downstream tasks. 
In this paper, we mainly focus on node attribute prompting scheme, i.e., Graph Prompt Feature (GPF)~\cite{GPF}, which aims to conduct prompting on graph node attributes. 
In this case, our GSP problem  is specifically defined as sparsely selecting some reliable 
node attributes for prompting. 
We show that this goal can be well formulated 
by 
introducing
some sparsity-inducing constraint/regularization into the objective of graph prompt optimization, as inspired by 
sparse representation theory~\cite{sparse_paper_1,sparse_paper_2} in machine learning field. 
Specifically,  
we propose two kinds of graph sparse prompting over attributes, termed Graph Sparse Feature Prompting (GSFP) and  Graph Sparse  multi-Feature Prompting  (GSmFP). 
%
Both GSFP and GSmFP 
provide a general framework for tuning any specific pre-trained GNNs, which can 
achieve attribute selection and
compact prompt learning simultaneously without requiring any additional trainable parameters.  
A simple yet effective algorithm has been designed for solving GSFP and GSmFP problem.

Overall, the main contributions of this paper are summarized as follows, 

\begin{itemize}
    \item 
We introduce a new prompt problem, termed graph sparse prompting, for reliable and compact graph  fine-tuning. 

    \item 

    We propose  some  sparsity-inducing
regularizations into the graph prompt optimization and 
develop two kinds of graph prompts, i.e., Graph Sparse Feature Prompting (GSFP) and Graph Sparse multi-Feature Prompting (GSmFP) for graph fine-tuning.  

\item 
We derive a simple yet effective algorithm to optimize the proposed GSFP and GSmFP. 
    
\end{itemize}
Experiments on 16 widely-used benchmark datasets validate the effectiveness and advantages of the proposed GSFP and GSmFP approaches.

\section{Related Works}

\subsection{Graph Pre-training}
By learning transferable representations from large-scale unlabeled data~\cite{pretrain_survey}, pre-training can enhance the generalization ability of GNNs and reduce the need for large annotated datasets. 
DGI~\cite{DGI} maximizes the mutual information between local node features and global graph embedding.
GraphMAE~\cite{GraphMAE} applies a masked autoencoder to reconstruct masked node features for unsupervised representation learning. 
EdgePred~\cite{GAE,GPPT} predicts the existence of edges between nodes to capture local structural patterns. 
GraphCL~\cite{GraphCL} performs contrastive learning by maximizing the consistency between different augmented views of the same graph. 
SimGRACE~\cite{SimGRACE} uses a GNN model along with its perturbed version as two encoders to generate two correlated representations for contrasting which can simplify the contrastive learning process.
However, most graph pre-training methods follow the `pre-training + fine-tuning' paradigm. 
Due to the gap between pre-training and fine-tuning stages, the fine-tuned model may perform suboptimally or even worse than training from scratch~\cite{negative}.

\subsection{Graph Prompt Learning}
Graph prompt learning~\cite{graph_prompt_survey} has recently emerged as a novel paradigm inspired by prompt learning in natural language processing (NLP) and computer vision (CV). Unlike the traditional `pre-training + fine-tuning' paradigm, graph prompt learning introduces prompts to guide the learning process while keeping the pre-trained model parameters frozen and only fine-tuning the prompts. There are two main types of approaches, i.e., task aligning prompting and graph data prompting. 

\textbf{Task Aligning Prompting.} To address the gap between different graph learning tasks, some works aim to create unified task templates to align the pre-training and downstream tasks. 
For example, 
GPPT~\cite{GPPT} uses link prediction as pre-training and uses prompt tokens to turn node classification into its pre-training task.
GraphPrompt\cite{GraphPrompt} constructs a graph similarity template to unify pre-training and downstream tasks. 
All-in-One~\cite{All-in-One} transforms both node-level and edge-level tasks into graph level-tasks. 
PRODIGY~\cite{PRODIGY} uses a neighbor matching task for pre-training and employs prompt graphs to express downstream tasks as pre-training tasks. 
OFA~\cite{OFA} unifies different tasks into the same node binary classification.

\textbf{Graph Data Prompting.} 
In addition to task aligning prompting, some works facilitate the transfer of pre-trained knowledge by adding learnable prompts into the model's input or hidden layer.
For example, 
VNT~\cite{VNT} adds learnable prompt nodes to the original node set and then feeds them into the pre-trained Graph Transformer for information aggregation. 
All-in-One\cite{All-in-One} designs prompt as graph and considers the connection between input graph and prompt graph. 
GPF~\cite{GPF} adds the same learnable prompt vector to each node feature during the input phase. GPF-plus~\cite{GPF} further assigns different prompts to each node feature via cross-attention module. GraphPrompt\cite{GraphPrompt} multiplies the output data of model's hidden layer with a learnable prompt vector to assist the readout operation. 
HetGPT\cite{HetGPT} introduces a small number of trainable parameters into the node features of the heterogeneous graph to generate prompted node features. 
HGPrompt~\cite{HGPrompt} introduces feature prompt and heterogeneity prompt to modify the node features in the readout operation and auxiliary aggregation operation.
MDGPT~\cite{MDGPT} utilizes dual prompts to modify the downstream input features, facilitating the transfer of pre-trained knowledge to the target domain.

\section{Preliminary}
In this section, we briefly introduce two kinds of graph prompt learning methods, i.e., Graph Prompt Feature (GPF)~\cite{GPF} and Graph Prompt Feature-Plus (GPF-plus)~\cite{GPF}. 
Given a pre-trained GNN model $f_{\Phi}$ with 
parameters $\Phi$  and an input graph $\mathcal{G}(\mathbf{X}, \mathbf{A})$, the data prompt-based fine-tuning process can be formulated as follows,
\begin{align}\label{EQ:pt}
	\{\theta^{*}, \tilde{\mathcal{G}}^*\} = \text{argmin} \,\,\, \mathcal{L}^{down}\big(f_{\Phi}(\tilde{\mathcal{G}}(\tilde{{\textbf{X}}}, \tilde{\textbf{A}})), \theta \big)  
\end{align}
where 
$\tilde{\mathcal{G}}(\tilde{{\textbf{X}}}, \tilde{{\textbf{A}}})$
denotes the prompted graph generated by employing some prompting functions to adjust node features or edge weights for the downstream task,
and $\theta$ denotes the learnable parameters in the downstream task, such as the task-specific classifier head.

One important way to obtain the prompted graph is to 
adjust graph node features by 
adding some learnable prompt vectors to the node features/attributes, which is called Graph Prompt Feature (GPF)~\cite{GPF}. 
To be specific, given an input graph $\mathcal{G}(\textbf{X}, \textbf{A})$, where $\textbf{X} \in \mathbb{R}^{n \times d}$ denotes the node feature matrix and $\textbf{A} \in \mathbb{R}^{n \times n}$ represents the adjacency matrix, GPF generates a prompted graph $\tilde{\mathcal{G}}( \tilde{\textbf{X}},\textbf{A})$
by adding a shared learnable prompt vector $\textbf{p} \in \mathbb{R}^d$ to each node features as, 
\begin{equation}\label{eq:GPF_modification}
\tilde{\textbf{X}} = \textbf{X} + \textbf{1} \textbf{p}^{T}
\end{equation}
where $\textbf{1}$ is a column vector of ones with dimension $n$.
Hence, the above formulation Eq.(1) for GPF can be specified as, 
\begin{align}\label{EQ:pt}
&\min_{\theta,\textbf{p}}\,\,\mathcal{L}^{down}\big(f_{\Phi}(\tilde{\mathcal{G}}(\tilde{\textbf{X}}, \textbf{A})), \theta \big)  \\
& s.t. 
\,\,\tilde{\textbf{X}} = \textbf{X} + \textbf{1}\textbf{p}^T
\nonumber
\end{align}
%

Beyond a single prompt vector, GPF-plus~\cite{GPF}
further assigns a prompt vector to each node by using $k$ independent prompt basis vectors $\textbf{P}=[ \textbf{p}_1, \textbf{p}_2, \ldots, \textbf{p}_k] \in \mathbb{R}^{d\times k}$. 
It employs a cross-attention mechanism to assign these prompt vectors to all nodes as, 
\begin{align}\label{eq:combined_formula}
\tilde{\textbf{X}}=\textbf{X}+\textbf{S}\textbf{P}^T ,\, \textrm{where}\,\,
\textbf{S}_{ij} =
\frac{\exp(\textbf{b}_j^T \textbf{x}_i)}{\sum_{l=1}^{k} \exp(\textbf{b}_l^T \textbf{x}_i)}
\end{align}
where 
$\textbf{B} = [ \textbf{b}_1, \textbf{b}_2, \ldots, \textbf{b}_k ] \in \mathbb{R}^{d\times k}$ are $k$ learnable linear projections. 
When $k=1$, GPF-plus degenerates to a `weighted' GPF. 
For convenience, in the following presentation, we include $\textbf{B}$ in the learnable parameters $\theta$. 
Therefore, the above formulation Eq.(1) for GPF-plus can be specified as, 
\begin{align}\label{EQ:pt}
&\min_{\theta,\textbf{P}}\,\,\mathcal{L}^{down}\big(f_{\Phi}(\tilde{\mathcal{G}}(\tilde{\textbf{X}},\textbf{A})), \theta \big)  \\
& s.t. \,\,\tilde{\textbf{X}}=\textbf{X}+\textbf{S}\textbf{P}^T
\nonumber
\end{align}
Note that, the optimal $\theta^*$ and $\textbf{P}^*$ 
can be learned by using the gradient descent algorithm. 




\section{Graph Sparse Feature Prompting}\label{sec:methodology}

\begin{figure}[!ht]

	\centering
	\includegraphics[width=0.48\textwidth]{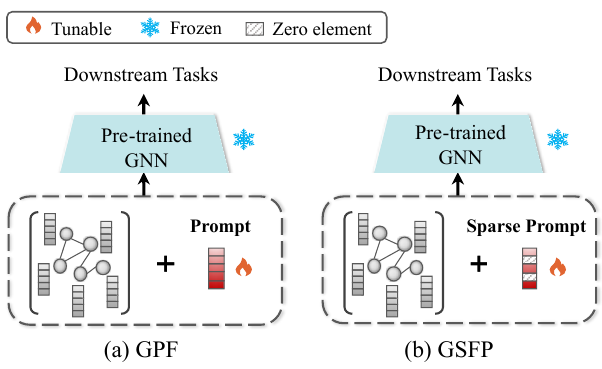}
	\caption{Comparing GPF and our method GSFP. GPF conducts prompting on every dimension of node features. In contrast, GSFP aims to conduct prompting on certain specific feature dimensions.}
 
 \label{fig:frame}
\end{figure}

As shown in Eqs.(3,5), the above GPFs adopt the deterministic prompting mechanism on all node attributes (features). 
This ‘complete’ prompting mechanism is obviously redundant and suboptimal. 
To overcome these issues, one
straightforward way is to select some reliable/meaningful
attributes for prompting, allowing the model to focus on some most relevant features. 
This can be achieved by adding a $\ell_0$-norm sparse constraint on prompt vector $\textbf{p}$ in Eq.(3) as,  
\begin{align}\label{EQ:pt}
&	 \min_{\theta,\textbf{p}}\,\,\mathcal{L}^{down}\big(f_{\Phi}(\tilde{\mathcal{G}}(\tilde{\textbf{X}},\textbf{A})), \theta \big)  \\
& s.t. \,\,\tilde{\textbf{X}} = \textbf{X}+\textbf{1}\textbf{p}^T, \|\textbf{p}\|_0\leq m \nonumber
\end{align} 
where $\|\textbf{p}\|_0$ denotes the number of non-zero elements in prompt vector $\textbf{p}$.
This constraint encourages the learned $\textbf{p}$ to be sparse, thereby naturally enabling the addition of prompt elements to be conducted only on some feature dimensions. 
In this paper, we call it as Graph Sparse Feature Prompting (GSFP).  
Since $\ell_0$-norm constraint is discrete and non-convex, it is difficult to optimize it directly. Similar to Lasso~\cite{Lasso}, we use $\ell_1$-norm regularization to approximate the above problem as follows,
\begin{align}\label{EQ:pt}
&\min_{\theta,\textbf{p}}\,\,\mathcal{L}^{down}\big(f_{\Phi}(\tilde{\mathcal{G}}(\tilde{\textbf{X}},\textbf{A})), \theta \big) +\lambda \|\textbf{p}\|_1 \\
& s.t. \,\,\tilde{\textbf{X}}=\textbf{X}+\textbf{1} \textbf{p}^T \nonumber
\end{align}
where $\lambda > 0$ is the penalty parameter and $\|\textbf{p}\|_{1}=\sum_{i}|\textbf{p}_i|$ denotes the $\ell_1$-norm function. Figure~\ref{fig:frame} shows the intuitive explanation of our GSFP model.  

\begin{figure}[!htp]  
    \centering
    \includegraphics[width=0.45\textwidth]{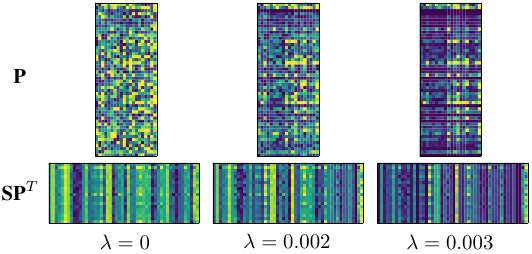}

    \caption{Visualizations of the learned matrix $\textbf{P} \in \mathbb{R}^{d\times k}$ and $\textbf{S}\textbf{P}^T \in \mathbb{R}^{n\times d}$ under different $\lambda$. We set the prompt basis vector number $k = 20$, feature dimension $d = 50$, and the node number $n = 20$ for demonstration. Here, we can note that, as the parameter $\lambda$ increases, there are more zero rows in $\textbf{P}$ and thus more zero columns in $\textbf{S}\textbf{P}^T$. }
       
    \label{fig:sparsity}
\end{figure}

The above GSFP can also be extended to the 
multiple prompts learning. 
Specifically, for the $k$-prompt formulation of GPF-plus in Eq.(5), 
we can encourage
prompt matrix $\textbf{P}$ to be row sparse in prompt learning, i.e., there are many zero rows in the learned $\textbf{P}$.
Note that, when $\textbf{P}$ is row sparse, the matrix $\textbf{S}\textbf{P}^T$ is column sparse (there are many zero columns), which thus allows the model to select
some meaningful attributes for prompting via Eq.(4). 
Therefore, we can extend GSFP to multi-prompt learning by 
simply adding a  $\ell_{2,1}$-norm sparse constraint~\cite{sparse_paper_1} on the prompt matrix $\textbf{P}$, i.e.,
\begin{align}\label{EQ:pt}
&	\min_{\theta,\textbf{P}}\,\,\mathcal{L}^{down}\big(f_{\Phi}(\tilde{\mathcal{G}}(\tilde{\textbf{X}},\textbf{A})), \theta \big)+\lambda\|\textbf{P}\|_{2,1}  \\
& s.t. \,\,\tilde{\textbf{X}}=\textbf{X}+\textbf{S}\textbf{P}^T\nonumber
\end{align}
where $\|\textbf{P}\|_{2,1} = \sum_{i}\sqrt{\sum_{j}|\textbf{P}_{ij}|^{2}}$ denotes the $\ell_{2,1}$-norm function and $\lambda > 0$ is the penalty parameter to control the row sparsity of learned $\textbf{P}$, i.e., larger $\lambda$ encourages more zero rows in $\textbf{P}$ and thus more zero columns in $\textbf{S}\textbf{P}^T$, as illustrated  in Figure~\ref{fig:sparsity}. 
In the following, we denote formulation Eq.(8) as 
 Graph Sparse multi-Feature Prompting (GSmFP). 

\section{Optimization}

In this section, we propose a simple yet effective algorithm to optimize the above GSFPs (GSFP, GSmFP). 
The proposed GSFPs involve $\ell_1$ and $\ell_{2,1}$ norm
regularization terms, which 
are difficult to employ 
gradient descent to optimize them directly. 
In this paper, we propose to utilize the Forward-backward splitting method~\cite{Forward_backward_splitting} to optimize them.


\subsection{Optimization of GSFP}

Our GSFP problem Eq.(7) has two learnable parameters, i.e., prompt $\textbf{p}$ and task head parameters $\theta$. In the following, we introduce the detailed optimization algorithms for them respectively. 

\textbf{Optimize $\textbf{p}$:} We adopt the Forward-backward splitting scheme to optimize prompt $\textbf{p}$ in GSFP problem Eq.(7). It conducts the optimization via the following update rule,
\begin{equation}
\textbf{p} \leftarrow \text{Prox1}_{\lambda} \left( \textbf{p} - \eta \nabla_{\textbf{p}} \mathcal{L}^{{down}} \right)
\end{equation}
where $\eta$ denotes the learning rate and $\lambda$ is the penalty parameter in Eq.(7). The proximal operator Prox1 is defined as follows,

\begin{equation}\label{eq:GSpPF_problem}
\text{Prox1}_{\lambda}(\textbf{y}) = \underset{\textbf{z}}{\arg \min}\,\, \frac{1}{2} \|\textbf{z} - \textbf{y}\|_2^2 + \lambda \|\textbf{z}\|_1
\end{equation}
It is known that problem Eq.(10) has the closed-form solution as,
\begin{equation}
\textbf{z}_i^* = \text{sign}(\textbf{y}_i) \max\big(|\textbf{y}_i| - \lambda, 0\big)
\end{equation}
where $\textbf{z}_i^*$ and $\textbf{y}_i$ represent the $i$-th element of vector $\textbf{z}^*$ and $\textbf{y}$, respectively.

\textbf{Optimize $\theta$:} The optimal $\theta^*$ is obtained by using gradient descent strategy, as commonly employed in many other works~\cite{GPF,All-in-One}.
The whole algorithm is summarized in Algorithm~\ref{algorithm:GSFP}.  

\subsection{Optimization of GSmFP}

\begin{figure}[!ht]
    \centering

    {    
        \begin{minipage}{0.5\linewidth}    
        \centering
        \includegraphics[width=1\linewidth]{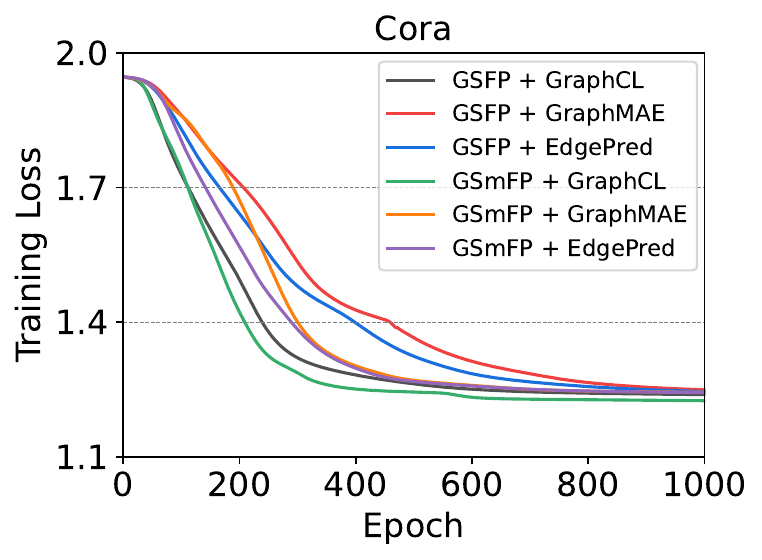}
        \end{minipage}
    }
    \hspace{-0.5cm} 
    {    
        \begin{minipage}{0.5\linewidth}    
        \centering
        \includegraphics[width=1\linewidth]{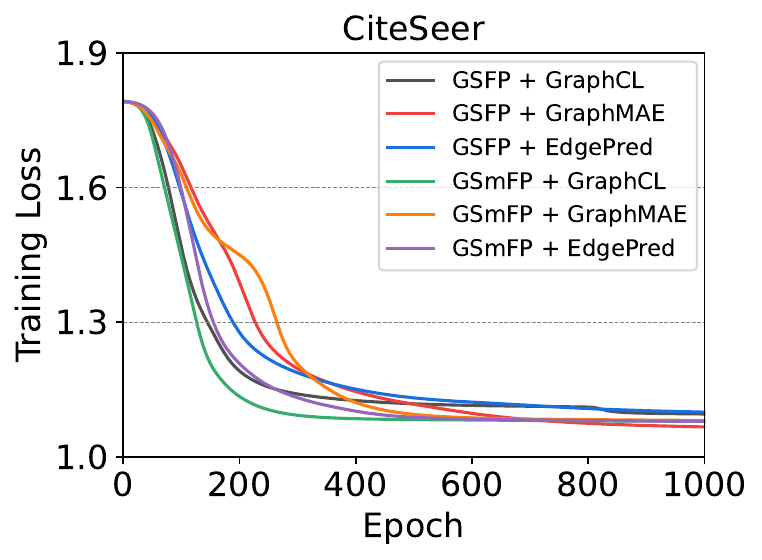}
        \end{minipage}
    }
    
    \caption{Training loss curves on Cora and CiteSeer datasets.}
    \label{fig:loss_curve}
\end{figure}

For GSmFP, we can also employ the Forward-backward splitting method to optimize prompt matrix $\textbf{P}$ and use gradient descent strategy to optimize $\theta$. 
Specifically, the update rule for optimizing $\textbf{P}$ is given  as,
\begin{equation}
\textbf{P} \leftarrow \text{Prox21}_{\lambda} \left(\textbf{P} - \eta \nabla_{\textbf{P}} \mathcal{L}^{{down}} \right)
\end{equation}
where $\eta$ is the learning rate and $\lambda$ is the penalty parameter in Eq.(8). The proximal operator Prox21 is defined as,
\begin{equation}\label{eq:GSpPF_plus_problem}
\text{Prox21}_{\lambda}(\textbf{Y}) = \underset{\textbf{Z}}{\arg \min}\,\, \frac{1}{2} \|\textbf{Z} - \textbf{Y}\|_F^2 + \lambda \|\textbf{Z}\|_{2,1}
\end{equation}
It is known that problem Eq.(13) has 
the closed-form solution which is given as, 
\begin{equation}
\textbf{Z}_i^* =
\begin{cases} 
\begin{alignedat}{2}
&\frac{\|\textbf{Y}_i\|_2 - \lambda}{\|\textbf{Y}_i\|_2} \textbf{Y}_i, \quad &&\text{if } \lambda < \|\textbf{Y}_i\|_2 \\
&\textbf{0}, \quad &&\text{otherwise}
\end{alignedat}
 \end{cases}
\end{equation}
where $\textbf{Z}_i^*$ and $\textbf{Y}_i$ represent the $i$-th row of the matrix $\textbf{Z}^*$ and $\textbf{Y}$, respectively. 
The whole algorithm is summarized in Algorithm~\ref{algorithm:GSmFP}. 
Figure~\ref{fig:loss_curve} shows the training loss curves of GSFP and GSmFP on Cora~\cite{citation_datasets} and CiteSeer~\cite{citation_datasets} datasets, demonstrating the convergence of the above Algorithm~\ref{algorithm:GSFP} and~\ref{algorithm:GSmFP}. 


\begin{algorithm}
	\caption{Optimization of GSFP}
        \label{algorithm:GSFP}
	\begin{algorithmic}
		\STATE \textbf{Input:} An input graph $\mathcal{G}(\textbf{X}, \textbf{A})$, learning rate $\eta$ and penalty parameter $\lambda$
		\STATE \textbf{Output:} Optimal prompt vector $\textbf{p}^*$ and parameters $\theta^{*}$
		\STATE Initialize $\textbf{p}$ and $\theta$
		\WHILE{not converged}
            \STATE $\textbf{p} \leftarrow \textbf{p} - \eta \nabla_{\textbf{p}} \mathcal{L}^{{down}}$
		\STATE $\textbf{p} \leftarrow \text{Prox1}_{\lambda}(\textbf{p})$
            \STATE $\theta \leftarrow \theta - \eta \nabla_{\theta} \mathcal{L}^{{down}}$
		\ENDWHILE
		\STATE \textbf{Return:} $\textbf{p}^*$, $\theta^{*}$
	\end{algorithmic}
\end{algorithm}


\begin{algorithm}
	\caption{Optimization of GSmFP}
        \label{algorithm:GSmFP}
	\begin{algorithmic}
		\STATE \textbf{Input:} An input graph $\mathcal{G}(\textbf{X}, \textbf{A})$, learning rate $\eta$ and penalty parameter $\lambda$
		\STATE \textbf{Output:} Optimal prompt matrix $\textbf{P}^*$ and parameters $\theta^{*}$
		\STATE Initialize $\textbf{P}$ and $\theta$
		\WHILE{not converged}
            \STATE $\textbf{P} \leftarrow \textbf{P} - \eta \nabla_{\textbf{P}} \mathcal{L}^{{down}}$
            \STATE $\textbf{P} \leftarrow \text{Prox21}_{\lambda} \left(\textbf{P} \right)$
            \STATE $\theta \leftarrow \theta - \eta \nabla_{\theta} \mathcal{L}^{{down}}$    		
		\ENDWHILE
		\STATE \textbf{Return:} $\textbf{P}^*$, $\theta^{*}$
	\end{algorithmic}
\end{algorithm}


\begin{table}[!ht]
    \centering
    \caption{Statistics of datasets.}
     \label{tab:datasets}
    \fontsize{7pt}{9pt}\selectfont 
    \begin{tabular}{>{\raggedright\arraybackslash}m{9em}>{\raggedright\arraybackslash}m{3em}>{\raggedright\arraybackslash}m{4em}>{\raggedright\arraybackslash}m{4em}>{\raggedright\arraybackslash}m{3em}>{\raggedright\arraybackslash}m{3em}>{\raggedright\arraybackslash}m{3em}}
        \toprule

        Dataset         & Graphs & Avg.nodes & Avg.edges & Features & Classes   \\

        \midrule

        Cora            & 1      & 2,708     & 10,556    & 1,433    & 7         \\
        CiteSeer        & 1      & 3,327     & 9,104     & 3,703    & 6         \\
        PubMed          & 1      & 19,717    & 88,648    & 500      & 3         \\
        Photo           & 1      & 7,650     & 238,162   & 745      & 8         \\
        Computers       & 1      & 13,752    & 491,722   & 767      & 10        \\
        CS              & 1      & 18,333    & 163,788   & 6,805    & 15        \\ 
        Physics         & 1      & 34,493    & 495,924   & 8,415    & 5         \\
        Wisconsin       & 1      & 251       & 515       & 1,703    & 5         \\
        Cornell         & 1      & 183       & 298       & 1,703    & 5         \\
        Chameleon       & 1      & 2,277     & 62,792    & 2,325    & 5         \\
        Ogbn-arxiv      & 1      & 169,343   & 1,166,243 & 128      & 40        \\

        \midrule

        COX2            & 467    & 41.22     & 43.45     & 3        & 2         \\
        MUTAG           & 188    & 17.93     & 19.79     & 7        & 2         \\
        IMDB-BINARY     & 1,000  & 19.77     & 96.53     & 0        & 2         \\
        IMDB-MULTI      & 1,500  & 13.00     & 65.94     & 0        & 3         \\
        REDDIT-MULTI-5K & 4,999  & 508.52    & 594.87    & 0        & 5         \\
        
        \bottomrule
    \end{tabular}
\end{table}


\section{Experiment}
\subsection{Experimental Setup}
\textbf{Datasets.}
To evaluate our proposed GSFP and GSmFP, we conduct experiments on 16 widely-used benchmark datasets on both node classification and graph classification tasks. 
For node classification task,  we use eleven datasets including three citation networks (Cora~\cite{citation_datasets}, CiteSeer~\cite{citation_datasets}, PubMed~\cite{citation_datasets}), two Amazon co-purchase datasets (Amazon Photo~\cite{copurchase_datasets}, Amazon Computers~\cite{copurchase_datasets}), two co-author networks (Coauthor CS~\cite{coauthor_datasets}, Coauthor Physics~\cite{coauthor_datasets}), three heterophilic web page networks (Wisconsin~\cite{webpage_datasets}, Cornell~\cite{webpage_datasets}, Chameleon~\cite{webpage_datasets}) and one large-scale dataset (Ogbn-arxiv~\cite{Ogbnarxiv_dataset}). 
For graph classification task, we use five datasets including two molecule datasets (COX2~\cite{COX2_dataset}, MUTAG~\cite{MUTAG_dataset}), two movie collaboration datasets (IMDB-BINARY~\cite{IMDB_REDDIT_datasets}, IMDB-MULTI~\cite{IMDB_REDDIT_datasets}) and one social network dataset (REDDIT-MULTI-5K~\cite{IMDB_REDDIT_datasets}).
The detailed descriptions of these datasets are summarized in Table~\ref{tab:datasets}.
Similar to works~\cite{ProG,GCOPE},  we adopt 1-shot setting for training and split the remaining samples into validation and test sets at a 1:9 ratio.

\textbf{Baselines.}
We compare our methods with three kinds of approaches: 
\textit{(1) Supervised method:} We choose GCN~\cite{GCN} as the backbone network under supervised setting for the comparison. 
\textit{(2) `Pre-training + fine-tuning' methods:} GraphMAE~\cite{GraphMAE} leverages a masked autoencoder to learn node representations by reconstructing the masked features.  EdgePred~\cite{GPPT} randomly drops a portion of edges and trains the GNN model to reconstruct them. GraphCL~\cite{GraphCL} employs contrastive learning to maximize the agreement between different augmented views of the same graph.
\textit{(3) Prompt learning methods:} We select several mainstream prompt learning methods including All-in-One~\cite{All-in-One}, GPF~\cite{GPF}, GPF-plus~\cite{GPF}, GraphPrompt~\cite{GraphPrompt} and GPPT~\cite{GPPT}. All-in-One concatenates the input graph with the prompt graph. Both GPF and GPF-plus add prompt vectors to the input features. GraphPrompt uses a prompt vector to assist the readout operation. GPPT transforms the node classification task into the link prediction pre-training via prompt tokens. 

\textbf{Implementation details.}
We conduct all experiments based on ProG library~\cite{ProG}. For a fair comparison, we use a 3-layer GCN~\cite{GCN} architecture with a hidden dimension of 256 for all methods.
For all pre-training methods, we pre-train the model on Flickr~\cite{Flickr_dataset} dataset and then fine-tune it on other node classification datasets. Therefore, we apply a learnable linear transformation to unify all feature dimensions to 100. 
Similarly, for graph classification task, we pre-train the model on DD~\cite{DD_datasets} dataset and then fine-tune it on other graph classification datasets.  

\textbf{Parameter setup.}
During the pre-training phase, the learning rate is set to 1e-3 and the weight decay is set to 1e-5. For downstream tasks, we set the learning rate to 1e-3 and the weight decay to 5e-4 for all methods. For other parameters in the comparison methods, we generally follow the default parameters in the original publications. 
For GPF-plus~\cite{GPF} and our GSmFP, $k$ is set to 10 and 20 respectively for node classification and graph classification task. Besides, we determine the optimal parameter $\lambda$ values based on the performance on validation set. 

\begin{table}[!ht]
\renewcommand{\thetable}{III}
\centering
\caption{Accuracy of 1-shot graph classification.}
\label{tab:graph_classification_result}
\fontsize{6pt}{11pt}\selectfont 

\begin{tabular}{>{\raggedright\arraybackslash}m{0.5em}|>{\raggedright\arraybackslash}m{4.5em}|*{5}{>{\raggedright\arraybackslash}m{4em}}}
\toprule

\multicolumn{2}{l|}{Method \textbackslash Dataset}           
                                          &COX2       &MUTAG      &IMDB-B     &IMDB-M      &RDT-M5K \\ 
\midrule

\multicolumn{2}{l|}{GCN}           
                                          &63.87±6.69 &55.95±13.59&50.08±0.72 &36.94±2.71 &21.99±3.02 \\
\midrule

\multirow{6}{*}{\rotatebox{90}{EdgePred}}    
                             & FT         &60.76±2.84 &53.10±17.28&55.68±7.19 &34.35±5.79 &22.53±2.86 \\
                             & All-in-One &60.43±7.59 &62.02±14.10&61.71±5.26 &36.22±5.97 &26.20±4.21 \\
                             & GPF        &60.81±4.03 &56.43±15.44&62.14±5.18 &37.18±6.46 &23.29±0.92 \\
                             & GPF-plus   &56.47±9.08 &57.98±11.48&60.87±4.71 &37.98±5.67 &24.48±3.19 \\
                             & GSFP       &64.73±7.69 &58.45±12.88&64.45±2.13 &38.04±4.63 &23.37±2.64 \\
                             & GSmFP      &62.20±10.06&58.69±10.84&62.65±6.74 &40.86±3.39 &25.78±7.47 \\
\midrule

\multirow{6}{*}{\rotatebox{90}{GraphCL}} 
                             & FT         &62.63±5.58 &55.71±13.55&50.06±0.72 &33.95±1.20 &22.09±2.40 \\
                             & All-in-One &57.85±7.20 &64.52±19.22&59.38±5.64 &36.57±4.38 &22.67±3.67 \\
                             & GPF        &63.48±4.96 &55.83±14.05&62.14±5.12 &37.77±5.44 &23.46±4.12 \\
                             & GPF-plus   &61.24±4.34 &53.81±17.65&61.74±5.70 &37.91±5.49 &25.86±4.77 \\
                             & GSFP       &63.34±6.94 &58.10±13.23&63.45±2.80 &38.19±5.31 &24.31±3.85 \\
                             & GSmFP      &63.58±7.01 &55.83±12.45&65.05±1.13 &39.07±6.09 &26.13±4.94 \\
\midrule

\multirow{6}{*}{\rotatebox{90}{GraphMAE}}    
                             & FT         &59.76±7.24 &49.76±15.85&52.41±6.01 &32.95±0.29 &22.60±2.73 \\
                             & All-in-One &55.32±27.44&65.71±25.80&61.47±3.28 &37.05±6.31 &22.81±3.61 \\
                             & GPF        &57.66±6.04 &59.64±14.80&62.16±4.92 &36.91±6.96 &23.39±0.90 \\
                             & GPF-plus   &64.15±7.37 &52.98±19.15&61.22±4.60 &37.33±6.94 &23.19±0.91 \\
                             & GSFP       &63.39±9.17 &59.29±14.01&63.36±3.89 &37.12±7.00 &23.91±2.01 \\
                             & GSmFP      &66.06±7.55 &58.93±14.19&62.45±3.51 &37.77±7.67 &23.01±0.74 \\
\midrule

\multicolumn{2}{l|}{GraphPrompt}           
                                          &61.19±6.48 &54.05±16.02&58.20±4.88 &34.84±3.87 &23.86±1.44 \\
\bottomrule

\end{tabular}
\end{table}
\setcounter{table}{1}
\begin{table*}[!ht]
\centering
\caption{Accuracy of 1-shot node classification.}
\label{tab:node_classification_result}
\fontsize{7pt}{11pt}\selectfont 

\begin{tabular}{>{\raggedright\arraybackslash}m{0.5em}|>{\raggedright\arraybackslash}m{4.5em}|*{11}{>{\raggedright\arraybackslash}m{4em}}}
\toprule

\multicolumn{2}{l|}{Method \textbackslash Dataset}           
                  &Cora       &CiteSeer   &PubMed     &Photo      &Computers  &CS         &Physics    &Wisconsin  &Cornell    &Chameleon  &Ogbn-arxiv\\
\midrule

\multicolumn{2}{l|}{GCN}           
                  &29.84±9.15 &27.63±6.36 &51.52±10.97&39.45±8.54 &34.94±11.03&57.43±7.86 &53.05±21.47&20.18±5.20 &23.98±6.50 &22.39±1.61 & 5.27±2.18 \\
\midrule

\multirow{6}{*}{\rotatebox{90}{EdgePred}}    
     & FT         &31.11±6.41 &26.92±4.20 &51.17±11.69&40.81±5.73 &33.47±11.21&55.49±7.13 &52.67±19.07&23.69±3.27 &22.24±6.77 &25.14±3.57 & 6.04±2.18 \\
     & All-in-One &20.44±8.12 &19.73±2.37 &42.08±7.82 &34.04±5.25 &27.84±6.73 &36.41±4.41 &43.22±9.15 &21.26±9.31 &16.15±7.20 &21.90±1.67 & 0.95±1.26 \\
     & GPF        &35.16±12.74&33.19±5.04 &48.14±7.67 &46.79±6.65 &39.58±11.52&64.89±8.79 &66.48±20.03&25.59±10.41&25.22±8.26 &22.61±2.46 &10.30±3.43 \\
     & GPF-plus   &34.04±11.06&33.46±4.60 &47.07±7.49 &47.52±5.66 &37.36±9.09 &62.96±9.25 &70.36±18.10&27.57±9.38 &24.72±8.65 &23.05±2.80 & 9.58±2.60 \\
     & GSFP       &36.07±13.60&33.20±6.82 &51.35±10.72&48.00±4.86 &42.16±9.30 &65.61±8.60 &73.70±19.15&25.50±10.19&25.47±8.63 &22.67±2.32 &11.17±3.81 \\
     & GSmFP      &36.03±13.14&33.52±4.76 &51.90±12.52&49.56±7.07 &43.21±8.93 &64.11±10.59&74.17±17.08&27.57±9.17 &24.97±8.08 &22.86±2.67 &10.24±2.69 \\
\midrule

\multirow{6}{*}{\rotatebox{90}{GraphCL}} 
     & FT         &30.89±8.66 &28.52±6.26 &48.80±7.42 &36.98±7.21 &31.69±5.00 &54.37±5.66 &64.65±14.07&20.54±12.53&26.21±7.62 &24.11±3.01 & 4.76±1.21 \\
     & All-in-One &14.70±7.43 &20.61±1.21 &41.45±6.49 &29.52±1.58 &24.34±5.63 &35.27±4.51 &43.80±8.16 &25.68±14.77&18.63±11.21&21.71±1.24 & 3.09±4.83 \\
     & GPF        &33.24±13.86&32.77±5.44 &48.25±7.84 &43.28±5.53 &34.52±8.45 &61.83±6.64 &69.59±16.99&26.04±9.18 &23.48±5.87 &23.70±3.58 & 6.24±1.52 \\
     & GPF-plus   &32.68±10.75&27.86±5.94 &48.21±8.21 &42.64±8.07 &33.46±7.10 &56.97±7.56 &69.83±15.56&26.04±8.66 &25.09±8.10 &23.61±3.21 & 7.38±2.14 \\
     & GSFP       &33.20±13.85&33.08±5.70 &53.56±10.36&49.56±8.73 &38.46±11.05&63.41±9.54 &74.62±14.92&27.57±8.14 &25.71±6.47 &25.45±3.94 & 9.35±2.49 \\
     & GSmFP      &34.08±13.43&33.60±5.75 &55.74±13.36&46.71±8.34 &36.60±7.89 &60.43±6.37 &72.83±16.55&25.68±9.20 &25.59±8.03 &26.81±1.30 & 9.12±2.17 \\
\midrule

\multirow{6}{*}{\rotatebox{90}{GraphMAE}}    
     & FT         &30.49±9.08 &31.40±5.29 &49.93±9.62 &40.54±11.22&32.79±10.00&54.99±7.46 &46.88±16.11&20.09±4.55 &24.47±7.09 &23.64±2.86 & 5.15±2.93 \\
     & All-in-One &12.54±2.52 &20.71±1.46 &39.82±9.14 &42.56±7.47 &31.32±9.30 &37.39±5.00 &39.74±10.10&24.14±13.95&20.62±10.51&21.65±1.45 & 3.09±4.83 \\
     & GPF        &35.15±12.96&32.77±5.86 &48.11±7.79 &46.44±7.37 &39.24±10.32&63.31±10.46&75.28±15.69&24.05±10.04&24.35±8.16 &22.55±2.44 & 9.11±2.40 \\
     & GPF-plus   &34.05±9.41 &31.05±5.50 &48.47±7.49 &46.41±6.39 &39.73±11.91&65.30±8.81 &69.71±17.24&27.03±9.31 &24.60±8.72 &23.55±3.42 & 7.72±1.23 \\
     & GSFP       &38.56±8.96 &34.61±6.90 &50.74±8.61 &49.53±4.58 &41.21±10.80&64.81±9.78 &75.32±15.57&25.14±9.83 &25.09±9.09 &22.57±2.46 &10.81±3.88 \\
     & GSmFP      &39.29±7.79 &33.61±5.30 &50.70±8.78 &49.99±5.75 &42.89±10.71&65.33±8.82 &75.37±17.64&27.12±9.77 &25.34±7.97 &25.16±2.98 &10.65±0.70 \\
\midrule

\multicolumn{2}{l|}{GPPT}           
                  &20.11±8.82 &26.07±6.36 &59.93±7.30 &58.62±9.27 &37.89±12.82&62.40±7.16 &84.91±3.34 &23.15±6.09 &22.98±9.61 &24.16±4.14 & 5.62±4.29 \\
\midrule

\multicolumn{2}{l|}{GraphPrompt}           
                  &34.88±10.64&25.67±4.40 &46.02±6.72 &47.10±6.83 &34.07±7.81 &61.03±10.52&69.02±4.90 &21.53±5.61 &24.97±8.06 &22.88±2.37 & 9.55±2.94 \\
\bottomrule

\end{tabular}
\end{table*}
\setcounter{table}{3}

\subsection{Comparison Results}
\textbf{Node classification.}
In Table~\ref{tab:node_classification_result}, we show the results of all methods with five random splits under the 1-shot setting. One can observe that  
(1) Compared to traditional `pre-training + fine-tuning' methods, most graph prompt learning methods obtain better results, which indicates that prompt learning can effectively mitigate the negative transfer issue caused by the task and graph domain gap between the pre-training and downstream stages.
(2) The proposed GSFP and GSmFP gain significant improvements over GPF~\cite{GPF} and GPF-plus~\cite{GPF} which also conduct prompting on node features. 
This validates the effectiveness of the proposed GSFPs which conduct prompting on specific node feature dimensions and thus perform more optimally and reliably.
(3) Our methods can obtain the best (or 2nd best) performance on all datasets, which further demonstrates the superiority of GSFP and GSmFP. 

\textbf{Graph classification.}
In Table~\ref{tab:graph_classification_result}, we show the results of all methods under the 1-shot setting on five graph classification datasets. One can note that the proposed GSFP and GSmFP perform better than GPF~\cite{GPF} and GPF-plus~\cite{GPF} in most cases, which is consistent with the results in Table~\ref{tab:node_classification_result}. 
Besides, our methods almost achieve the best results on all datasets, which further shows the effectiveness of the proposed methods on graph classification task.

\begin{figure*}[!htp]
    \centering
    {    
        \begin{minipage}{0.25\linewidth}    
        \centering
        \includegraphics[width=1\linewidth]{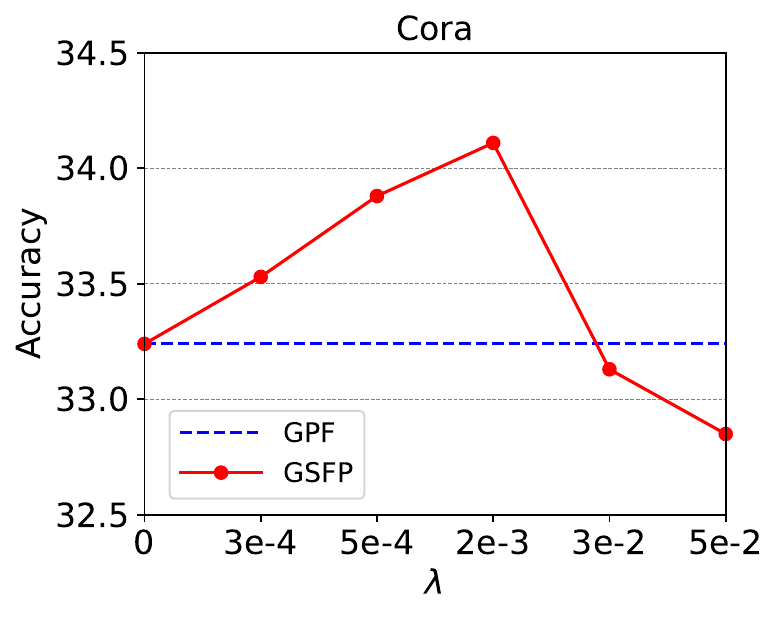}
        \end{minipage}
    }
    \hspace{-0.5cm} 
    {    
        \begin{minipage}{0.25\linewidth}    
        \centering
        \includegraphics[width=1\linewidth]{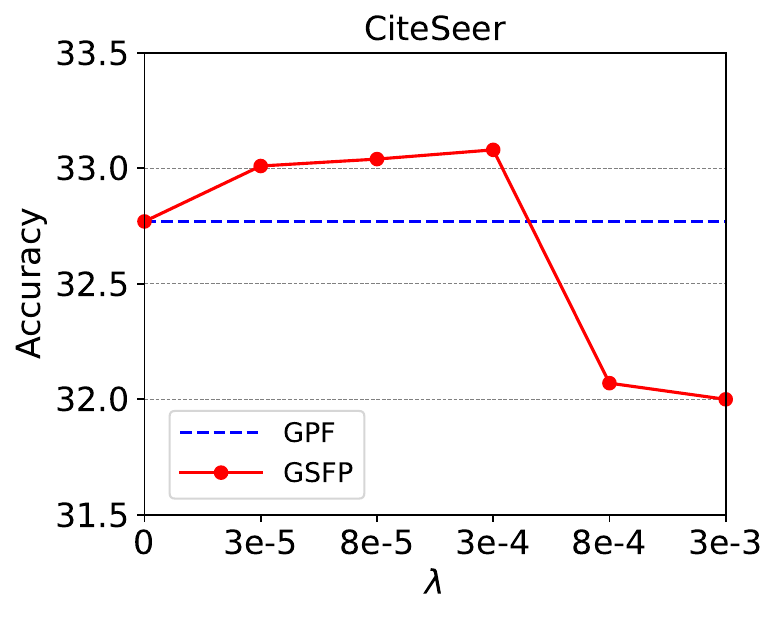}
        \end{minipage}
    }
    \hspace{-0.5cm} 
    {    
        \begin{minipage}{0.25\linewidth}    
        \centering
        \includegraphics[width=1\linewidth]{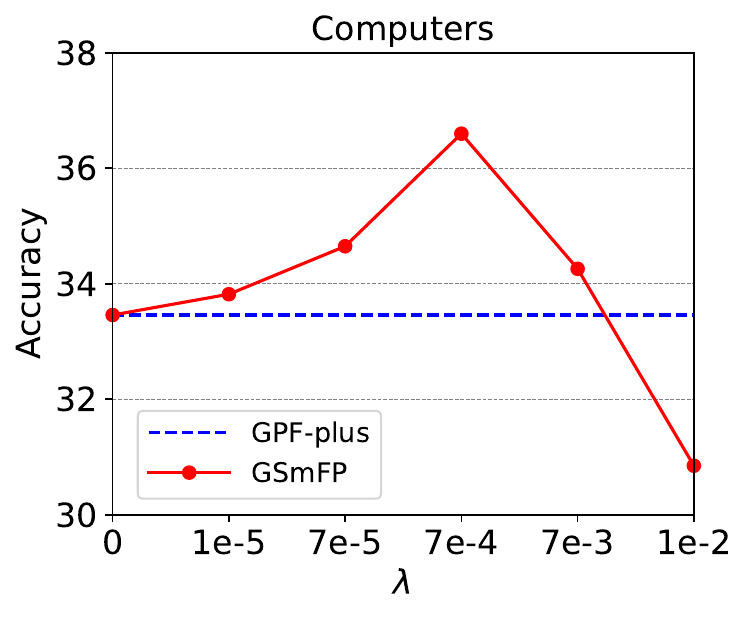}
        \end{minipage}
    }
    \hspace{-0.5cm} 
    {    
        \begin{minipage}{0.25\linewidth}    
        \centering
        \includegraphics[width=1\linewidth]{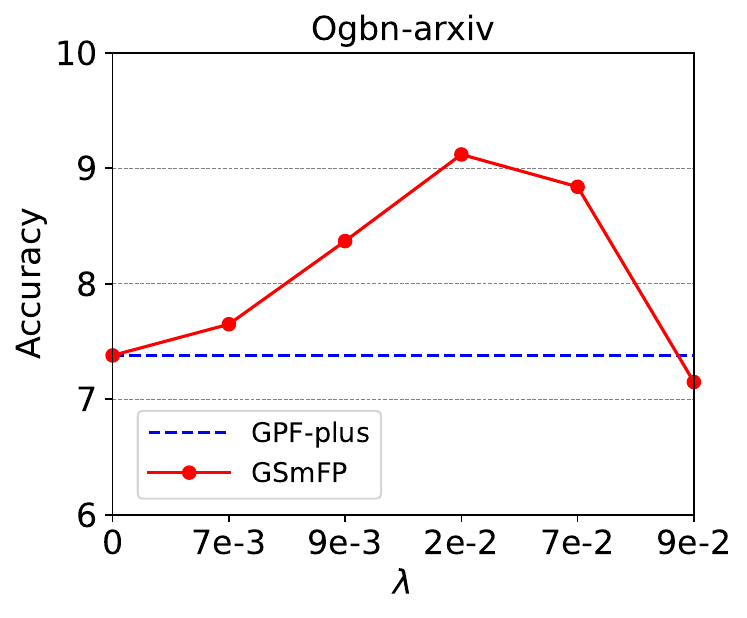}
        \end{minipage}
    }

    \vspace{-0.1cm} 
    
    {    
        \begin{minipage}{0.25\linewidth}    
        \centering
        \includegraphics[width=1\linewidth]{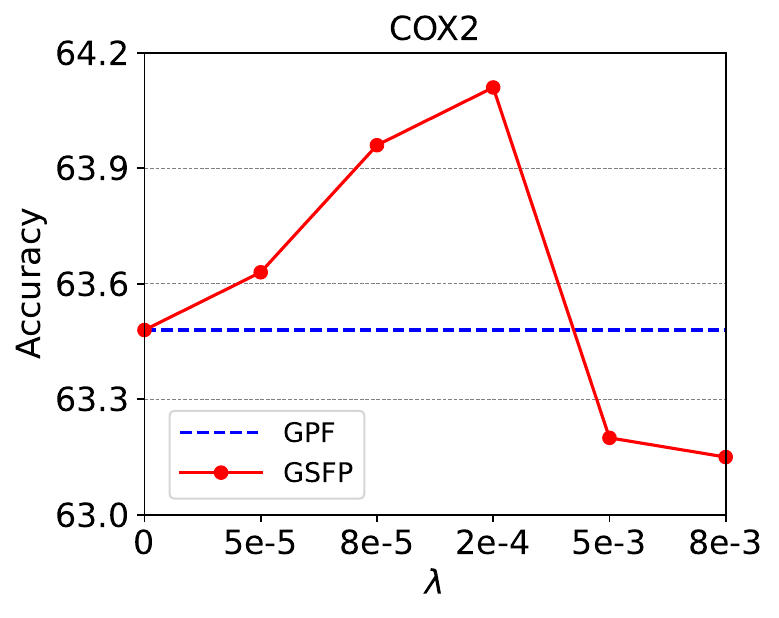}
        \end{minipage}
    }
    \hspace{-0.5cm} 
    {    
        \begin{minipage}{0.25\linewidth}    
        \centering
        \includegraphics[width=1\linewidth]{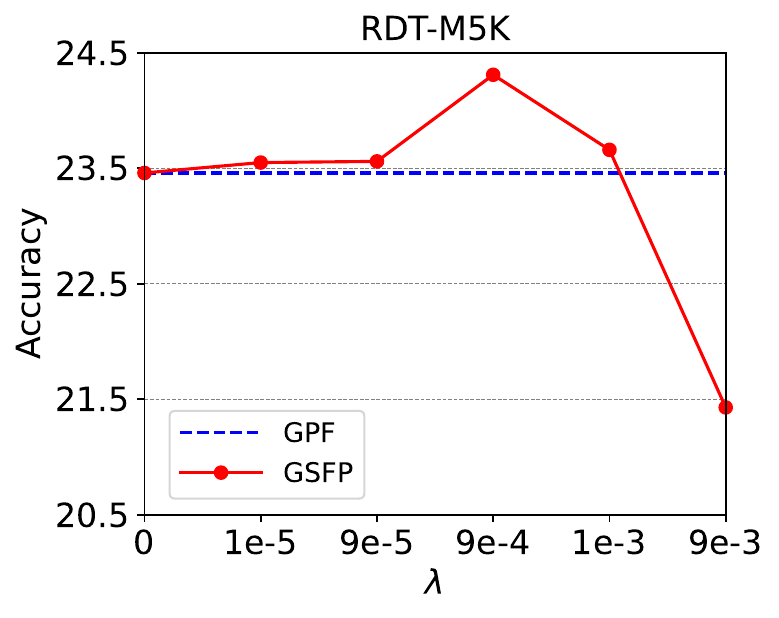}
        \end{minipage}
    }
    \hspace{-0.5cm} 
    {    
        \begin{minipage}{0.25\linewidth}    
        \centering
        \includegraphics[width=1\linewidth]{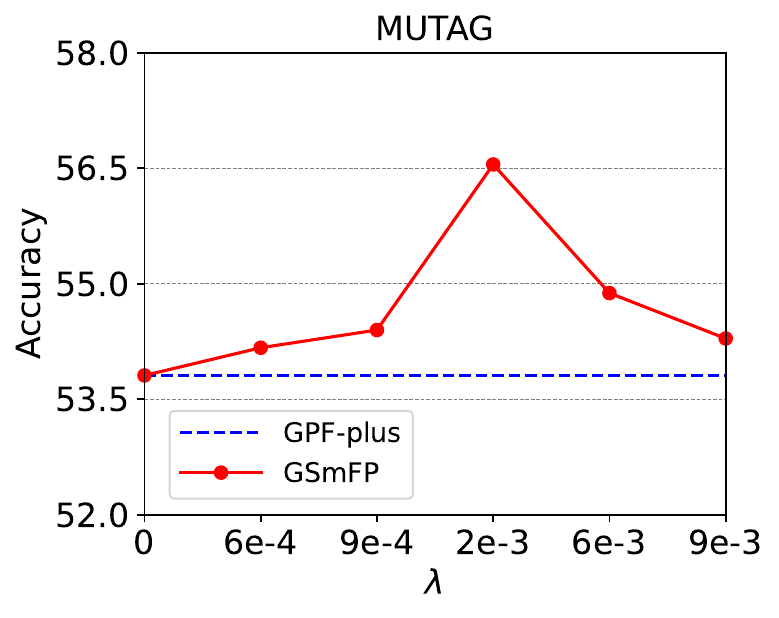}
        \end{minipage}
    }
    \hspace{-0.5cm} 
    {    
        \begin{minipage}{0.25\linewidth}    
        \centering
        \includegraphics[width=1\linewidth]{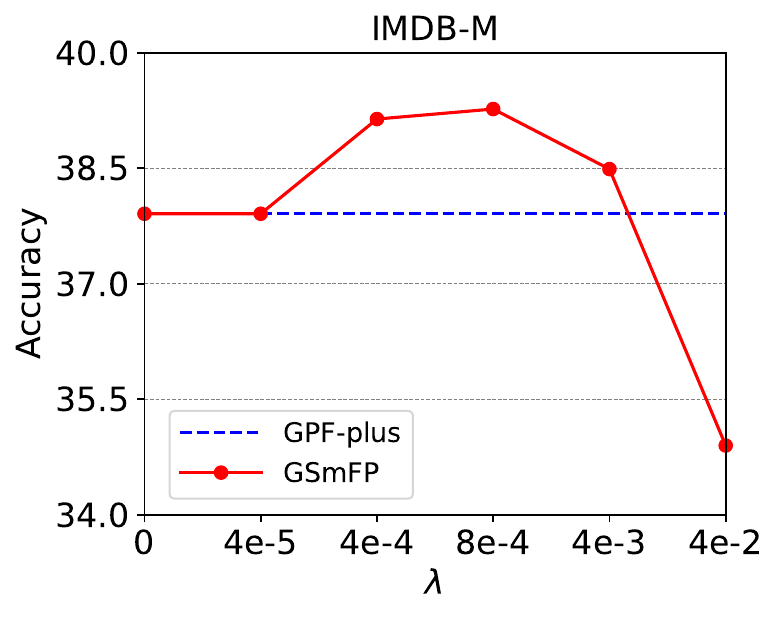}
        \end{minipage}
    }
    \caption{Analyze the impact of parameter $\lambda$ on different datasets.}
    \label{fig:lambda}
\end{figure*}

\subsection{Parameter Analysis}
As mentioned in Section~\ref{sec:methodology}, the parameter $\lambda$ controls the sparsity of the learned prompt. 
Thus, we analyze the effect of different $\lambda$ values on performance. Note that when $\lambda = 0$, the proposed GSFP and GSmFP degrade to GPF~\cite{GPF} and GPF-plus~\cite{GPF}, respectively. 
As shown in Figure~\ref{fig:lambda}, one can see that our methods achieve better performance than the baseline ($\lambda=0$) within a certain range. 
Besides, as $\lambda$ continues to increase, prompting can only be conducted on very few or no feature dimensions, which leads to a performance drop. 
This demonstrates the effectiveness of conducting prompting selectively on specific feature dimensions. 

\begin{figure}[!htp]
    \centering

    {    
        \begin{minipage}{0.5\linewidth}    
        \centering
        \includegraphics[width=1\linewidth]{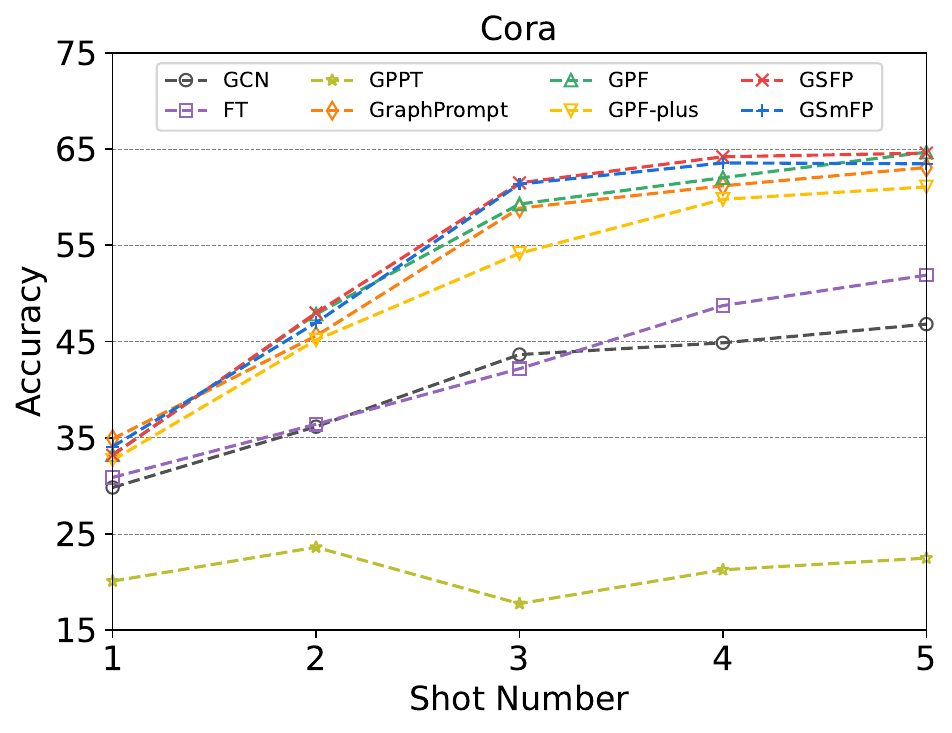}
        \end{minipage}
    }
    \hspace{-0.5cm} 
    {    
        \begin{minipage}{0.5\linewidth}    
        \centering
        \includegraphics[width=1\linewidth]{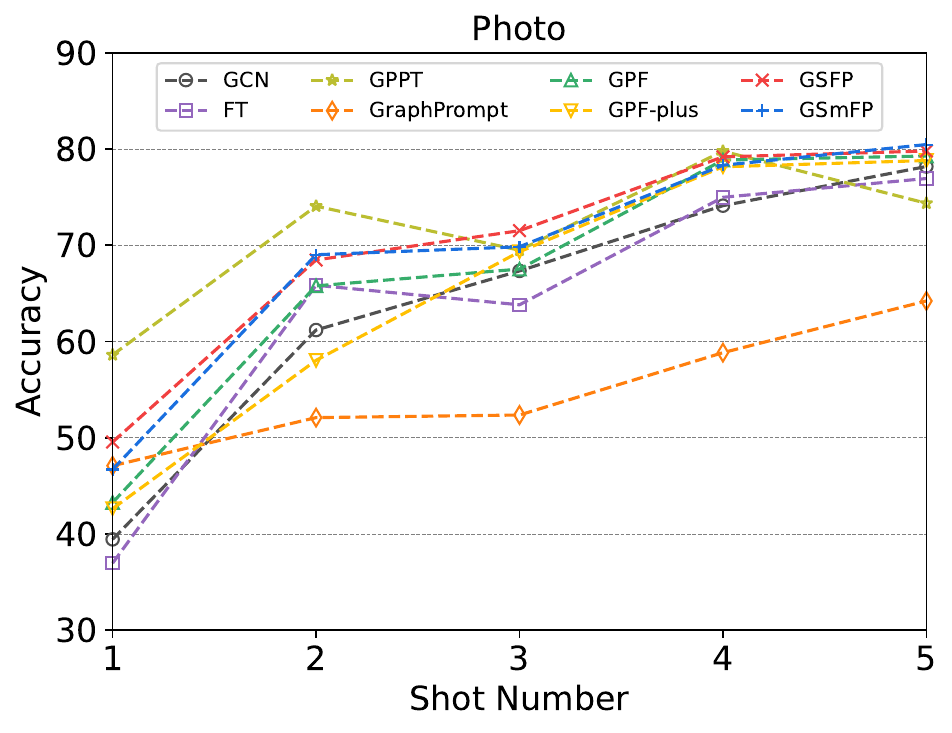}
        \end{minipage}
    }
    
    \caption{Analyze the impact of different shot numbers on Cora and Photo datasets.}
    \label{fig:shot_number}
\end{figure}


\begin{table}[!ht]
\centering
\caption{Performance on different backbones.}
\label{tab:different_backbones_result}
\fontsize{7pt}{11pt}\selectfont 

\begin{tabular}{m{0.5em}| l | m{4em} m{4em} m{4em} m{4em}}

\toprule

\multicolumn{2}{l|}{Dataset} 
                                             & \multicolumn{2}{l}{Cora}  & \multicolumn{2}{l}{Photo} \\
\cmidrule(lr){1-2} \cmidrule(lr){3-4} \cmidrule(lr){5-6}

\multicolumn{2}{l|}{Method \textbackslash Backbone} 
                                             & GAT         & SAGE        & GAT         & SAGE \\
\midrule

\multicolumn{2}{l|}{Supervised}          
                                             &32.26±7.99 &22.06±6.10 &37.08±8.20 &38.78±10.02\\
\midrule

\multirow{6}{*}{\rotatebox{90}{GraphCL}} 
                                & FT         &33.92±7.33 &30.30±7.82 &30.75±11.67&34.28±5.24 \\
                                & All-in-One &21.97±11.92&37.49±9.25 &26.14±5.45 &29.60±6.72 \\
                                & GPF        &33.32±9.14 &34.01±9.97 &36.71±4.51 &37.06±2.48 \\
                                & GPF-plus   &36.25±6.62 &32.47±6.95 &33.37±2.93 &34.51±8.19 \\
                                & GSFP       &36.64±10.03&35.77±10.17&38.57±11.16&37.93±3.47 \\
                                & GSmFP      &36.67±6.29 &33.23±6.51 &38.01±5.60 &39.54±5.02 \\
\midrule

\multirow{6}{*}{\rotatebox{90}{GraphMAE}}    
                                & FT         &35.70±9.54 &20.68±7.43 &36.32±7.15 &35.09±4.06 \\
                                & All-in-One &19.98±8.44 &15.28±3.81 &26.48±5.05 &39.95±6.37 \\
                                & GPF        &35.18±11.20&37.71±6.15 &33.91±5.08 &46.13±4.83 \\
                                & GPF-plus   &34.32±8.35 &36.65±10.55&37.06±6.10 &45.55±4.62 \\
                                & GSFP       &38.79±9.22 &38.00±5.91 &43.31±6.74 &47.67±4.28 \\
                                & GSmFP      &37.32±8.16 &36.68±10.52&40.95±8.80 &48.08±5.55 \\
\midrule
  
\multicolumn{2}{l|}{GPPT}        
                                             &31.04±15.07&20.67±4.06 &48.48±4.00 &53.86±10.73\\
\midrule

\multicolumn{2}{l|}{GraphPrompt}  
                                             &37.93±12.16&35.95±9.68 &47.61±3.64 &48.08±11.85\\
\bottomrule

\end{tabular}
\end{table}


%
\subsection{Additional Experiments}
\textbf{Impact of different shot numbers.} 
Figure~\ref{fig:shot_number} shows the performance across different shot numbers on Cora~\cite{citation_datasets} and Photo~\cite{copurchase_datasets} datasets. 
One can observe that most methods show improved performance with the increasing shot number. Our GSFP and GSmFP obtain a better advantage in most cases.

\textbf{Performance on different backbones.} 
Table~\ref{tab:different_backbones_result} shows the results of using GAT~\cite{GAT} and GraphSAGE~\cite{GraphSAGE} as backbones on Cora~\cite{citation_datasets} and Photo~\cite{copurchase_datasets} datasets.
Here, GSFP and GSmFP obtain improvements compared to GPF~\cite{GPF} and GPF-plus~\cite{GPF}, and they are also competitive with other comparison methods. 
This further demonstrates the generality and advantage of our GSFP methods on different GNN backbones. 



\section{Conclusion}

This paper first arises a new graph prompting problem, termed Graph Sparse Prompting (GSP), which is defined as adaptively and sparsely selecting the optimal graph
elements (this paper focuses on node attributes) to achieve compact prompting for downstream tasks. 
Then, 
we propose two kinds of graph sparse prompting over attributes, termed GSFP and GSmFP, by leveraging 
sparsity-inducing constraint/regularization into the graph prompt optimization objective. We derive an effective
algorithm to optimize them. 
Experiments on many benchmark datasets demonstrate that the proposed GSFPs outperform many other
comparison baselines on several pre-trained GNNs and different downstream tasks. 

\bibliography{main}

\begin{thebibliography}{10}

\bibitem{social_networks_gnn}
X.~Li, L.~Sun, M.~Ling, and Y.~Peng, ``A survey of graph neural network based
  recommendation in social networks,'' {\em Neurocomputing}, vol.~549,
  p.~126441, 2023.

\bibitem{drug_discovery_gnn}
J.~Xiong, Z.~Xiong, K.~Chen, H.~Jiang, and M.~Zheng, ``Graph neural networks
  for automated de novo drug design,'' {\em Drug Discovery Today}, vol.~26,
  no.~6, pp.~1382--1393, 2021.

\bibitem{fraud_detection_gnn}
Y.~Liu, X.~Ao, Z.~Qin, J.~Chi, J.~Feng, H.~Yang, and Q.~He, ``Pick and choose:
  A gnn-based imbalanced learning approach for fraud detection,'' in {\em
  Proceedings of the Web Conference 2021}, p.~3168–3177, 2021.

\bibitem{traffic_predictio_gnn}
L.~Kong, H.~Yang, W.~Li, J.~Guan, and S.~Zhou, ``Traffexplainer: A framework
  towards gnn-based interpretable traffic prediction,'' {\em IEEE Transactions
  on Artificial Intelligence}, pp.~1--15, 2024.

\bibitem{pretrain_survey}
Y.~Xie, Z.~Xu, J.~Zhang, Z.~Wang, and S.~Ji, ``Self-supervised learning of
  graph neural networks: A unified review,'' {\em IEEE Transactions on Pattern
  Analysis and Machine Intelligence}, vol.~45, no.~2, pp.~2412--2429, 2023.

\bibitem{negative_transfer}
L.~Wang, M.~Zhang, Z.~Jia, Q.~Li, C.~Bao, K.~Ma, J.~Zhu, and Y.~Zhong, ``Afec:
  Active forgetting of negative transfer in continual learning,'' {\em Advances
  in Neural Information Processing Systems}, vol.~34, pp.~22379--22391, 2021.

\bibitem{graph_prompt_survey}
X.~Sun, J.~Zhang, X.~Wu, H.~Cheng, Y.~Xiong, and J.~Li, ``Graph prompt
  learning: A comprehensive survey and beyond,'' {\em arXiv:2311.16534}, 2023.

\bibitem{GPPT}
M.~Sun, K.~Zhou, X.~He, Y.~Wang, and X.~Wang, ``Gppt: Graph pre-training and
  prompt tuning to generalize graph neural networks,'' in {\em Proceedings of
  the 28th ACM SIGKDD Conference on Knowledge Discovery and Data Mining},
  p.~1717–1727, 2022.

\bibitem{GraphPrompt}
Z.~Liu, X.~Yu, Y.~Fang, and X.~Zhang, ``Graphprompt: Unifying pre-training and
  downstream tasks for graph neural networks,'' in {\em Proceedings of the ACM
  Web Conference 2023}, p.~417–428, 2023.

\bibitem{All-in-One}
X.~Sun, H.~Cheng, J.~Li, B.~Liu, and J.~Guan, ``All in one: Multi-task
  prompting for graph neural networks,'' in {\em Proceedings of the 29th ACM
  SIGKDD Conference on Knowledge Discovery and Data Mining}, p.~2120–2131,
  2023.

\bibitem{AAGOD}
Y.~Guo, C.~Yang, Y.~Chen, J.~Liu, C.~Shi, and J.~Du, ``A data-centric framework
  to endow graph neural networks with out-of-distribution detection ability,''
  in {\em Proceedings of the 29th ACM SIGKDD Conference on Knowledge Discovery
  and Data Mining}, p.~638–648, 2023.

\bibitem{PSP}
Q.~Ge, Z.~Zhao, Y.~Liu, A.~Cheng, X.~Li, S.~Wang, and D.~Yin, ``Psp:
  Pre-training and structure prompt tuning for graph neural networks,'' in {\em
  Joint European Conference on Machine Learning and Knowledge Discovery in
  Databases}, p.~423–439, 2024.

\bibitem{GPF}
T.~Fang, Y.~Zhang, Y.~YANG, C.~Wang, and L.~Chen, ``Universal prompt tuning for
  graph neural networks,'' in {\em Advances in Neural Information Processing
  Systems}, vol.~36, pp.~52464--52489, 2023.

\bibitem{SUPT}
J.~Lee, W.~Yang, and J.~Kang, ``Subgraph-level universal prompt tuning,'' {\em
  arXiv preprint arXiv:2402.10380}, 2024.

\bibitem{GSPF}
B.~Jiang, H.~Wu, Z.~Zhang, B.~Wang, and J.~Tang, ``A unified graph selective
  prompt learning for graph neural networks,'' {\em arXiv preprint
  arXiv:2406.10498}, 2024.

\bibitem{VNT}
Z.~Tan, R.~Guo, K.~Ding, and H.~Liu, ``Virtual node tuning for few-shot node
  classification,'' in {\em Proceedings of the 29th ACM SIGKDD Conference on
  Knowledge Discovery and Data Mining}, p.~2177–2188, 2023.

\bibitem{theoretical_support}
Q.~Wang, X.~Sun, and H.~Cheng, ``Does graph prompt work? a data operation
  perspective with theoretical analysis,'' {\em arXiv preprint
  arXiv:2410.01635}, 2024.

\bibitem{sparse_paper_1}
F.~Nie, H.~Huang, X.~Cai, and C.~Ding, ``Efficient and robust feature selection
  via joint \(\ell_{2,1}\)-norms minimization,'' in {\em Advances in Neural
  Information Processing Systems}, vol.~23, pp.~1813--1821, 2010.

\bibitem{sparse_paper_2}
J.~Wright, A.~Y. Yang, A.~Ganesh, S.~S. Sastry, and Y.~Ma, ``Robust face
  recognition via sparse representation,'' {\em IEEE Transactions on Pattern
  Analysis and Machine Intelligence}, vol.~31, no.~2, pp.~210--227, 2009.

\bibitem{DGI}
P.~Veličković, W.~Fedus, W.~L. Hamilton, P.~Liò, Y.~Bengio, and R.~D. Hjelm,
  ``Deep graph infomax,'' in {\em International Conference on Learning
  Representations}, 2019.

\bibitem{GraphMAE}
Z.~Hou, X.~Liu, Y.~Cen, Y.~Dong, H.~Yang, C.~Wang, and J.~Tang, ``Graphmae:
  Self-supervised masked graph autoencoders,'' in {\em Proceedings of the 28th
  ACM SIGKDD Conference on Knowledge Discovery and Data Mining}, p.~594–604,
  2022.

\bibitem{GAE}
T.~N. Kipf and M.~Welling, ``Variational graph auto-encoders,'' {\em NeurIPS
  Workshop on Bayesian Deep Learning}, 2016.

\bibitem{GraphCL}
Y.~You, T.~Chen, Y.~Sui, T.~Chen, Z.~Wang, and Y.~Shen, ``Graph contrastive
  learning with augmentations,'' in {\em Advances in Neural Information
  Processing Systems}, vol.~33, pp.~5812--5823, 2020.

\bibitem{SimGRACE}
J.~Xia, L.~Wu, J.~Chen, B.~Hu, and S.~Z. Li, ``Simgrace: A simple framework for
  graph contrastive learning without data augmentation,'' in {\em Proceedings
  of the ACM Web Conference 2022}, p.~1070–1079, 2022.

\bibitem{negative}
W.~Jin, T.~Derr, H.~Liu, Y.~Wang, S.~Wang, Z.~Liu, and J.~Tang,
  ``Self-supervised learning on graphs: Deep insights and new direction,'' {\em
  arXiv preprint arXiv:2006.10141}, 2020.

\bibitem{PRODIGY}
Q.~Huang, H.~Ren, P.~Chen, G.~Kr\v{z}manc, D.~Zeng, P.~S. Liang, and
  J.~Leskovec, ``Prodigy: Enabling in-context learning over graphs,'' in {\em
  Advances in Neural Information Processing Systems}, vol.~36,
  pp.~16302--16317, 2023.

\bibitem{OFA}
H.~Liu, J.~Feng, L.~Kong, N.~Liang, D.~Tao, Y.~Chen, and M.~Zhang, ``One for
  all: Towards training one graph model for all classification tasks,'' in {\em
  The Twelfth International Conference on Learning Representations}, 2024.

\bibitem{HetGPT}
Y.~Ma, N.~Yan, J.~Li, M.~Mortazavi, and N.~V. Chawla, ``Hetgpt: Harnessing the
  power of prompt tuning in pre-trained heterogeneous graph neural networks,''
  in {\em Proceedings of the ACM Web Conference 2024}, p.~1015–1023, 2024.

\bibitem{HGPrompt}
X.~Yu, Y.~Fang, Z.~Liu, and X.~Zhang, ``Hgprompt: Bridging homogeneous and
  heterogeneous graphs for few-shot prompt learning,'' in {\em Proceedings of
  the AAAI Conference on Artificial Intelligence}, vol.~38, pp.~16578--16586,
  2024.

\bibitem{MDGPT}
X.~Yu, C.~Zhou, Y.~Fang, and X.~Zhang, ``Text-free multi-domain graph
  pre-training: Toward graph foundation models,'' {\em arXiv preprint
  arXiv:2405.13934}, 2024.

\bibitem{Lasso}
R.~Tibshirani, ``Regression shrinkage and selection via the lasso,'' {\em
  Journal of the Royal Statistical Society Series B: Statistical Methodology},
  vol.~58, no.~1, pp.~267--288, 1996.

\bibitem{Forward_backward_splitting}
P.~L. Combettes and J.-C. Pesquet, ``Proximal splitting methods in signal
  processing,'' {\em Fixed-Point Algorithms for Inverse Problems in Science and
  Engineering}, pp.~185--212, 2011.

\bibitem{citation_datasets}
P.~Sen, G.~Namata, M.~Bilgic, L.~Getoor, B.~Galligher, and T.~Eliassi-Rad,
  ``Collective classification in network data,'' {\em AI magazine}, vol.~29,
  no.~3, pp.~93--93, 2008.

\bibitem{copurchase_datasets}
O.~Shchur, M.~Mumme, A.~Bojchevski, and S.~G{\"u}nnemann, ``Pitfalls of graph
  neural network evaluation,'' {\em NeurIPS Workshop on Relational
  Representation Learning}, 2018.

\bibitem{coauthor_datasets}
A.~Sinha, Z.~Shen, Y.~Song, H.~Ma, D.~Eide, B.-J.~P. Hsu, and K.~Wang, ``An
  overview of microsoft academic service (mas) and applications,'' in {\em
  Proceedings of the 24th International Conference on World Wide Web},
  p.~243–246, 2015.

\bibitem{webpage_datasets}
H.~Pei, B.~Wei, K.~C.-C. Chang, Y.~Lei, and B.~Yang, ``Geom-gcn: Geometric
  graph convolutional networks,'' in {\em International Conference on Learning
  Representations}, 2020.

\bibitem{Ogbnarxiv_dataset}
W.~Hu, M.~Fey, M.~Zitnik, Y.~Dong, H.~Ren, B.~Liu, M.~Catasta, and J.~Leskovec,
  ``Open graph benchmark: Datasets for machine learning on graphs,'' in {\em
  Advances in Neural Information Processing Systems}, vol.~33,
  pp.~22118--22133, 2020.

\bibitem{COX2_dataset}
R.~A. Rossi and N.~K. Ahmed, ``The network data repository with interactive
  graph analytics and visualization,'' in {\em Proceedings of the Twenty-Ninth
  AAAI Conference on Artificial Intelligence}, p.~4292–4293, 2015.

\bibitem{MUTAG_dataset}
N.~Kriege and P.~Mutzel, ``Subgraph matching kernels for attributed graphs,''
  in {\em Proceedings of the 29th International Coference on International
  Conference on Machine Learning}, p.~291–298, 2012.

\bibitem{IMDB_REDDIT_datasets}
P.~Yanardag and S.~Vishwanathan, ``Deep graph kernels,'' in {\em Proceedings of
  the 21th ACM SIGKDD International Conference on Knowledge Discovery and Data
  Mining}, p.~1365–1374, 2015.

\bibitem{ProG}
C.~Zi, H.~Zhao, X.~Sun, Y.~Lin, H.~Cheng, and J.~Li, ``Prog: A graph prompt
  learning benchmark,'' {\em Advances in Neural Information Processing
  Systems}, 2024.

\bibitem{GCOPE}
H.~Zhao, A.~Chen, X.~Sun, H.~Cheng, and J.~Li, ``All in one and one for all: A
  simple yet effective method towards cross-domain graph pretraining,''
  p.~4443–4454, 2024.

\bibitem{GCN}
T.~N. Kipf and M.~Welling, ``Semi-supervised classification with graph
  convolutional networks,'' in {\em International Conference on Learning
  Representations}, 2017.

\bibitem{Flickr_dataset}
H.~Zeng, H.~Zhou, A.~Srivastava, R.~Kannan, and V.~Prasanna, ``Graphsaint:
  Graph sampling based inductive learning method,'' in {\em International
  Conference on Learning Representations}, 2020.

\bibitem{DD_datasets}
P.~D. Dobson and A.~J. Doig, ``Distinguishing enzyme structures from
  non-enzymes without alignments,'' {\em Journal of Molecular Biology},
  vol.~330, no.~4, pp.~771--783, 2003.

\bibitem{GAT}
P.~Veličković, G.~Cucurull, A.~Casanova, A.~Romero, P.~Liò, and Y.~Bengio,
  ``Graph attention networks,'' in {\em International Conference on Learning
  Representations}, 2018.

\bibitem{GraphSAGE}
W.~Hamilton, Z.~Ying, and J.~Leskovec, ``Inductive representation learning on
  large graphs,'' in {\em Advances in Neural Information Processing Systems},
  vol.~30, pp.~1024--1034, 2017.

\end{thebibliography}
\bibliographystyle{ieeetr}

\end{document}